\documentclass[3p,twocolumn]{elsarticle}
\usepackage{lineno,hyperref}
\usepackage[section]{placeins}
\usepackage[usenames, dvipsnames]{color}
\usepackage[labelfont=bf]{caption}
\usepackage{amsmath}

\usepackage[ruled,noline]{algorithm2e}
\usepackage{amssymb}
\usepackage{subfigure}
\usepackage{gensymb}
\usepackage{float}

\usepackage{multirow}
\usepackage{array}

\usepackage{chngcntr}

\newcommand{\tcr}[1]{\textcolor{red}{#1}}

\modulolinenumbers[5]
\begin{document}
\begin{frontmatter}

\title{Analysis of planar ornament patterns via motif asymmetry assumption and local connections}

\author{V.~Adanova}
\ead{venera@ceng.metu.edu.tr}

\author{S.~Tari \corref{cor1}}
\ead{stari@metu.edu.tr}

\cortext[cor1]{Corresponding author}
\address{Department of Computer Engineering, Middle East Technical University, 06800 Ankara}

\begin{abstract}
Planar ornaments,  a.k.a. wallpapers, are regular repetitive patterns which exhibit translational symmetry in two independent directions. There are exactly $17$ distinct planar symmetry groups.  We present a fully automatic method for complete analysis of  planar ornaments  in $13$ of these groups, specifically,  the groups called $p6m, \, p6, \, p4g, \,p4m, \,p4, \, p31m, \,p3m, \, p3,   \, cmm, \, pgg, \, pg, \, p2$ and $p1$. Given the image of an ornament fragment, we present a method to simultaneously classify the input into one of the $13$ groups and extract the  so called {\sl fundamental domain} (FD),  the minimum region that is sufficient to reconstruct the entire ornament. 
A nice feature of our method is that even when the given ornament image is a small portion such that it does not contain multiple translational units, the symmetry group as well as the fundamental domain can still be defined. This is because, in contrast to common approach, we do not attempt to first identify a global translational repetition  lattice. Though the presented  constructions work for quite a wide range of ornament patterns, a key assumption we make is that the perceivable motifs (shapes that repeat) alone do not provide clues for  the underlying symmetries of the ornament. In this sense, our main target is the planar arrangements of asymmetric interlocking shapes, as in the symmetry art of Escher.
 \end{abstract}

\end{frontmatter}

\section{Introduction}

\noindent

\begin{figure}[!hbt]
\centering
\fbox{
 \begin{tabular}{c}
\includegraphics[width=0.8\linewidth]{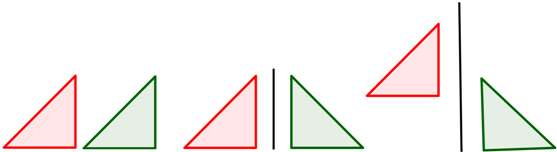} \\
\hspace{-0.1cm}(a)  \hspace{1.5cm} (b)  \hspace{1.5cm}(c)\\
\\
\fbox{\includegraphics[width=0.6\linewidth]{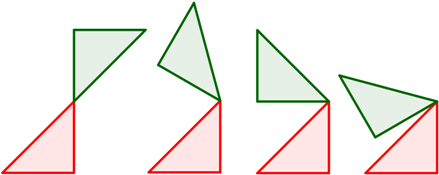}} 
\\
  \hspace{0.3cm} (d) 
\end{tabular}
}
\caption{4 primitive operations.(a) Translation, (b) reflection, (c) glide reflection, (d) $4$ types of rotational symmetry. }
\label{fig:operations}
\end{figure}

\noindent  Planar ornaments,  a.k.a. wallpapers, are repetitive patterns which exhibit translational symmetry in two independent directions. 
They form a tiling of the plane. They are created by repeating base unit in a predictable manner, using four primitive planar geometric operations: translation, rotation, reflection and glide reflection (Fig.~\ref{fig:operations}). Using combinations of these primitive operations applied on a base unit, different patterns can be generated. 
An interesting fact is that the four primitive operations can be combined in  exactly seventeen different ways to tile a plane, forming the so called $17$ $Plane$ $Symmetry$ $Groups$  \cite{Polya}.


\begin{figure}[!htb]
\centering

\includegraphics[width=0.9\linewidth,trim={0 0 0 0},clip]{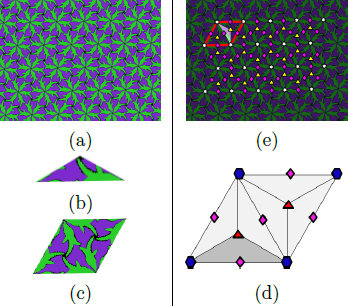} 
\caption{An example. (a) The ornament image, (b) fundamental domain (FD), (c) unit cell (UC), (d) symmetries and the relation between FD and UC, (e) symmetries, UC and FD superimposed on the ornament image.}
\label{fig:p6_example}
\end{figure}

We present an illustrative  example in Fig.~\ref{fig:p6_example}. 
Firstly, observe that the pattern Fig.~\ref{fig:p6_example} (a) can be generated by 
replicating the  equilateral triangular fragment depicted  in  Fig.~\ref{fig:p6_example}(b)  by  $60 \degree$ rotations in a systematic manner. This equilateral triangle fragment is the smallest fragment of the pattern that is sufficient to construct the entire pattern in Fig.~\ref{fig:p6_example}(a)  using four primitive operations. As such, it is referred as the  {\sl fundamental domain} (FD). 
Secondly, by rotating the FD $120 \degree$ twice around its top corner and then rotating $180 \degree$  around the middle point of its base yields the rhombus depicted in Fig.~\ref{fig:p6_example}(c). This is the smallest translational unit; the pattern in (a) can be generated by simply translating it along two independent directions; hence, it is referred as 
the {\sl unit cell} (UC). The abstracted unit cell of the tile, the relation between the fundamental domain and the unit cell along with the symmetries, is shown in Fig.~\ref{fig:p6_example} (d). The blue hexagons on the four corners of the rhombus indicate six-fold rotation centers,  and the red triangles and the  pink diamonds   located on the centers and the side mid-points of the two triangles forming the rhombus respectively indicate three-fold and two-fold rotation centers.  In the final illustration (Fig.~\ref{fig:p6_example}(e))  a sample cell is shown as superimposed on the original ornament image.

\begin{figure*}[!hbt]
\centering
\fbox{\includegraphics[width=14cm, height=15cm, keepaspectratio]{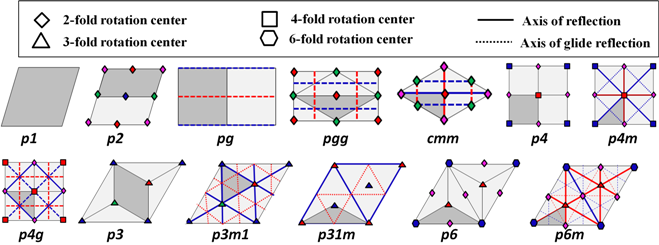}}\\
\caption{Unit cell structures for 13 Wallpaper Groups. Darker regions indicate fundamental domains.}
\hspace{1em}
\label{fig:cell_struct}
\end{figure*}

Through a common naming provided by {\sl {Crystallographic}} notation, the presented example happens to belong to a group called as $p6$, indicating six fold rotations. In the Crystallographic notation, the remaining $16$ groups are named as  $p1$, $pm$, $pg$, $cm$, $p2$, $pmm$, $pmg$, $pgg$, $cmm$, $p3$, $p3m1$, $p31m$, $p4$, $p4m$, $p4g$ and $p6m$.  In Fig.~\ref{fig:cell_struct}, we depict the  cell structures of each of the $13$ symmetry groups that we are interested in.

In the group name, each {character position} defines a  group property: 
The first position is either the letter $p$ which stands for $primitive$ $cell$ or  the letter $c$ which stands for $centered$ $cell$. 
The primitive cell is a unit cell with the centers of highest order of rotation at the vertices. The centered cell is encountered only in two cases ($cm$ and $cmm$ symmetry groups), and is chosen so that the reflection axis is normal to one or both sides of the cell. 
The digit  that follows the letters $p$ or $c$  indicates  the highest order of rotation, whereas 
the letter characters  $m$ and $g$ respectively stand for  mirror and glide reflections.  When there are two positions containing either of  $m$ and $g$, it is understood that the first reflection is normal to $x$ axis and the second is at an angle $\alpha$.
The digit denoting the highest order of rotation in a symmetry group can take only values 1, 2, 3, 4, and 6. This restriction is introduced by the $crystallographic$ $restriction$ $theorem$, which states that the patterns repeating in two dimension can only exhibit $180^\circ$, $120^\circ$, $90^\circ$, and $60^\circ$ rotations. 
\begin{figure}[htb]
\centering
  \begin{tabular}{m{2cm}m{2cm}m{2cm}}
\multicolumn{3}{c} {\includegraphics[width=0.9\linewidth]{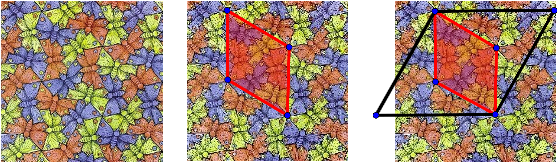} }\\
\hspace{0.8cm}(a) & \hspace{0.7cm} (b) & \hspace{1cm}(c)
\end{tabular}
\caption{An example case for a pattern with few repetitions. (a) Original ornament, (b) fundamental domain, (c) unit cell (black quadrilateral). Observe that the ornament fragment does not contain the full unit cell.}
\label{fig:smallcase}
\end{figure}

In the present paper, given an image fragment from an ornament belonging to either
of  the mentioned $13$ groups, we present a robust method to extract the fundamental domain  along with the underlying symmetry operations. That is, we provide a complete analysis of an ornamental pattern in $13$ groups based on an image.  
An important feature of our computational scheme is even when the given input image is only a  small portion  that a full unit cell does not fit to image (e.g., Fig.~\ref{fig:smallcase}(a)), the symmetry group as well as the fundamental domain can still be defined. Of course, once the symmetry group and fundamental domain are defined, it becomes trivial to deduce the translational unit cell.  We remark that most  of the existing methods rely on first discovering the underlying lattice via a translational repetition structure. This requires  global calculations, e.g., autocorrelations. In contrast, we do not search for a translational repetition or lattice. We directly look for local connections among motifs (protiles) from which we deduce symmetry clues that are later integrated via a decision tree.

\begin{figure*}[bt]
\centering
\includegraphics[width=0.93\linewidth, trim={0 0 0 0}, clip]{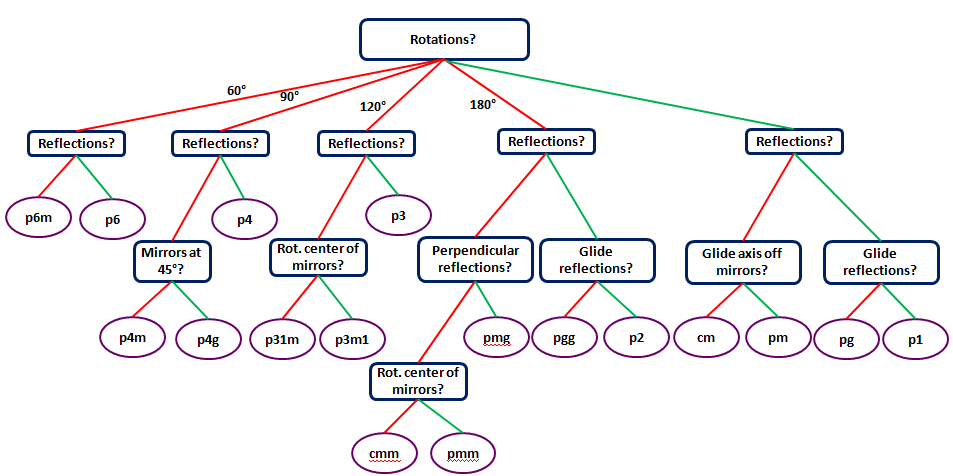}\\
\caption{Classical decision tree.}
\label{fig:decision_tree_classic}
\end{figure*}

Using a decision tree is indeed the classical method for grouping tiles into  symmetry groups based on individual clues \cite{Doris}; the classical decision tree is depicted in Fig.~\ref{fig:decision_tree_classic}; 
observe  that the check for mirror reflections dominate the yes/no question set. The rational behind the given question sequence could be  that it is easy for humans to spot mirror reflection and rotational symmetries while much harder to spot glides. While this is true, the robustness in the computational case might not be the same. 

We propose an alternative decision tree, of which details are given in \S~\ref{sec:method}. 
Initially, we accumulate indirect clues to mirror reflection as opposed to  searching for mirror reflection axis. We  postpone mirror reflection  control   till the last stage, and then at the last stage use mirror reflection check about the predicted axes only to eliminate possible false alarms resulting from indirect clues.
That is, we use mirror reflection check only to eliminate false alarms, not to catch missed ones. 
This means that we are willing to sacrifice mirror reflections in order not to wrongly assume that the tile has a mirror reflection. 
Our rational is as follows: If we only miss the mirror reflection, say classify a $p6m$ tile as $p6$, we still get the correct unit cell and a redundant (twice the size) fundamental domain. Hence, the pattern can be correctly generated.  If, however, we falsely classify a $p6$ tile as $p6m$, then the extracted fundamental domain is not sufficient, as we falsely assume the existence of a mirror reflection.
A consequence of not searching for mirror reflection is that we can only recognize groups that contain sufficient symmetries other than mirror. This leaves out $4$ of the groups, $pm, pmm, cm, pmg$ and gives us the $13$ groups listed above. 
For the $13$ groups, our goal is to obtain the fundamental domain robustly up to a mirror reflection. \\
In accumulating symmetry clues, we only resort to local connections among motifs (protiles). 
Many of the classical ornaments, such as those found in Islamic art, are constructed from symmetric protiles such as stars that provide a clue to the symmetry group of the tile. In some works, the symmetries of the protiles themselves are used as a clue to the symmetry. However, we believe that inferring symmetries of the motifs are not robust due to possible noise in motif extraction.
 Moreover, the ornament artist may be using nearly symmetric motifs  that do not reflect the symmetries of the ornament itself (several such examples exist in Escher's art). 
Hence, we neither attempt  to recover individual  motifs correctly nor check their symmetries. 
Indeed, we even expect the patterns we analyze  contain sufficient number of asymmetric motifs or at least the most significant symmetries are not completely swallowed by motifs having the same symmetry.

\begin{figure}[htb]
\centering
  \begin{tabular}{c}   
\includegraphics[width=0.94\linewidth,trim={0 0 0 0},clip]{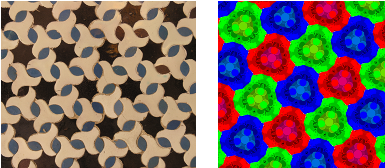}  
\end{tabular}
\caption{Possible valid inputs. }
\label{fig:tiles_example}
\end{figure}

Motif asymmetry assumption is not as restrictive as it may strike at first. 
Consider two valid examples shown in Fig.~\ref{fig:tiles_example}.  The first one is of group $p6$ and contains exactly three all symmetric protiles -- brown, beige and gray--  respectively having six-fold, three-fold and two-fold rotational symmetries. That is, the ornament does not contain any asymmetric motif and all three types of rotation centers coincide with the motif centers. Nevertheless, the local relations among the two-fold rotational symmetric gray motifs reveal both the six-fold and three-fold rotational centers, providing sufficient clues for $p6$. The second ornament, at a first glance, might give the impression that it contains only a single  form that has all the symmetries of the ornament itself, {\sl i.e.}, three-fold rotational symmetry and mirror reflection. Nevertheless, due to texture, there are also other detectable motifs in the form of circles or circle fragments, of which local relations provide clues for various three-fold rotation centers. \\
In general, our constructions can not handle ornaments of single motif such that the motif  contains all the symmetries of the symmetry group; for example, a checkerboard pattern or a uniform pack of triangles.  
We remark, however, that such ornaments are typically the most obvious ones to recognize. Furthermore, even for those class of ornaments, it is possible to identify the translation grid though full analysis revealing FD can not be performed. 

The rest of the paper is organized as follows. \S~\ref{sec:related_work} is on related work. The Method details are given in \S~\ref{sec:method}  and the results on $100$ tile set is in \S~\ref{sec:results}. Finally, \S~\ref{sec:conclusion} is the Summary and Conclusion.
\section{Related Work}
\label{sec:related_work}
\noindent Ornament patterns have always been a source of curiosity and interest, not only in arts and crafts but also other fields including mathematics, computation, cultural studies etc. Early researchers  mostly examined ornaments in cultural contexts, {\sl e.g.},  \cite{WashburnBook}, with a goal of  revealing social structures and their interaction via dominant symmetries used in the ornament designs of individual cultures or geographical regions. 
%
%
In mathematics, the ornament patterns are studied in terms of the groups formed by the symmetry operations. Few examples include \cite{Grunbaum86,Senechal88,Coxeter86}.
 The Dutch artist, Escher took a particular interest in patterns formed by repeating asymmetric shapes and discovered a local structure leading to the same wallpaper patterns; his work on symmetry is examined in \cite{DorisBook}. 
Regular repetitive patterns such as Wallpaper and Frieze groups are even utilized in quite practical problems; for example, to analyze human gait \cite{Liu_Gait} or to achieve automatic fabric defect detection in 2D patterned textures~\cite{Ngan2008,Asha2012,Ngan2010}. 

In the general pool of works in computational symmetry, the main focus has been finding symmetry axes in single objects.
Since a single object can exhibit only mirror reflections and rotational symmetries, the efforts are heavily  focused on reflections and rotations~\cite{Sun97,Sun99,Keller,Loy,Prasad2004,Prasad2005,Lee2008,Lee2010} or finding local symmetries a.k.a shape skeletons. To our knowledge, ~\cite{Liu2010_2,Lee2012} are the only works that address finding a glide reflection axis in an image, though the goal is to study one dimensional arrangements of symmetry, {\sl e.g.}, leaves. The works targeting  shape symmetry, whether directly from an image or from a segmented region, fall out of our focus. Our focus is on the symmetries of planar patterns formed by regular repetition of shapes via four primitive geometric transformations.\\

In analyzing planar periodic patterns, 
 translational symmetry, the most primitive repetition operation, is encountered in several works on recurring structure discovery~\cite{LiuJ2013,Doubek2009,Doubek2010,Torii2013,Gao2009}. 
The general flow of such works is to detect visual words and cluster them based on their appearance and spatial layout. Among these works, \cite{Doubek2009,Doubek2010,Torii2013} further perform image retrieval based on the discovered recurring structures. 
In~\cite{Doubek2009,Doubek2010} instead of directly using the recurring structures for image matching, the authors first detect a translational repetition lattice of an image. There can be multiple lattices for an image. Thus, given a query image with various detected lattices, they search a database for images with equivalent lattices. In each search, the matching score between two lattices is a product of two measurements: the similarity of the grayscale mean of the representative unit cell and the similarity of the color histograms. 
In~\cite{Park2008,Park2009} detection of deformed lattice in a given pattern is proposed. They first propose a seed lattice from detected interest points. Using those interest points a commonly occurring 
lattice vectors are extracted. Subsequently, the seed lattice is refined, and grown outward until it covers  whole pattern. 

In \cite{Han2014}, for translational symmetry, model based lattice estimation is performed where the model comparison for hypotheses generated via peaks of the autocorrelation is implemented using approximate marginal likelihood.

Some works move further beyond computing  a translational lattice to address 
classifying repeated patterns according to the 17 plane symmetry groups (wallpapers)~\cite{Liu98,Liu2000_1,Liu2000_5,Liu2004_pami,Liu2001}.
As a first step,  lattice detection is performed. 
After the lattice detection step, sequence of yes/no questions are answered, until the final symmetry group is determined. Since the possible lattice types that can be associated with are restricted by the symmetries the ornament pattern  exhibits, detecting the lattice types reduces the number of symmetry groups.
Though the lattice detection  is commonly performed using peak heights in autocorrelation \cite{Liu98}, in \cite{Liu2000_1}
an alternative peak detection algorithm based on so called {\sl regions of dominance} is used to detect patterns translational lattice.  The region of dominance is defined as the largest circle centered on the candidate peak such that no higher peaks are contained in the circle.  The authors argue  that the region of dominance is more important than the height of the peak.
A Hough transform is used to detect two shortest translation vectors that best explains the majority of the point data. In order to test whether a pattern has certain symmetry, the conjectured symmetry is applied to entire pattern and then the similarity between the original image and the transformed one is computed. The representative motif is chosen to be the most symmetrical figure. 

Recently, \cite{Albert2015}, combined lattice extraction and point symmetry groups of individual motifs to analyse Islamic patterns in mosaics. This method specifically targets Islamic ornaments in which motifs such as n-stars typically provide clues to the underlying plane symmetry group. As such, it is not readily applicable if motifs can not be robustly extracted or motifs do not reflect the symmetries. In, \cite{Nasri2015}, rotation groups are detected to analyze Islamic rosette patterns.

It is also possible to perform a continuous characterization of the ornament by comparing ornament images. This is for example encountered in \cite{AgustiMelchor2013}, 
 where ornament images are classified according to a symmetry feature vector calculated based on a prior lattice extraction and yes/no questions; 
 for lattice detection, they used method in \cite{AgustiMelchor2008}. In \cite{Adanova2016}, ornament images are directly compared in a transformed domain after applying a global transformation. 
 
Note that among the works addressing planar patterns, there are also several interesting works on pattern synthesis, including how to generate an ornament in a certain symmetry group, how to use a given motif to tile the plane in a certain style, or how to map a given wallpaper pattern to a curved surface \cite{Kaplan,vonGagern2009}.

\section{The Method}
\label{sec:method}
\noindent Our symmetry detection system has three modules: image processing module, local connectivity analysis, and final symmetry detection. Each of them are separately explained below.
\subsection{Image Processing} \label{ssec:IP}
\noindent The input to the image processing module is an arbitrary ornament $I$  which may be a noisy scanned image  or screen shot of a part of an ornament drawn using a computer tool. That is,  input ornament images are acquired in  arbitrary imaging conditions. 
The processing proceeds in three stages: gamma correction, initial clustering, and refinement.  At the end of the refinement step,  $k$ binary images for each ornament are obtained. {\sl These binary images will be called masks for that ornament image.}

The number of masks is a result of adaptive clustering. Quite often, the number of masks coincides with the number of colors in that ornament image. This, however, is not always so because the image processing module is not given the number of colors.  \\

\noindent {\bf{Gamma Correction.}} The first step of the image processing module  is gamma-correction ~\cite{Dijk2006}  to brighten the black lines and shadows. It is performed on the $Y$ component of $xyY$ color space using the following formula:
\[ Y_{out}=Y_{min}+(Y_{max}-Y_{min})*(\frac{Y-Y_{min}}{Y_{max}-Y_{min}})^\gamma\]

Resulting image depends on the $\gamma$ parameter. If one chooses $\gamma<1$ then the lightness of an image is higher than in original image and darker colors have more contrast. When $\gamma=1$ no effect on original image is observed, while $\gamma>1$ makes colors darker than in original image.  \\

\noindent {\bf{Initial Clustering.}} The next step is an iterative application of a clustering algorithm till the number of initial clusters drop below a pre-defined value $N_c$. Here, we assume that the number of distinct colors is less than $N_c$; hence, if the number of colors is more, then some color groups are merged to form bigger motif groups. For clustering, we use fast and robust mean shift algorithm \cite{Fukunaga1975}. It produces clusters based on a given feature space. In our case, the features are $L*a*b*$ channels of the gamma corrected image. The appealing feature of this algorithm for us is that it does not require the number of clusters to be specificied. (This is unlike k-means). However, it requires a bandwidth parameter which indirectly influence the number of detected clusters. 
We automate the clustering process by iteratively using mean shift clustering increasing the bandwidth at each iteration as follows. 

At the initial step, the bandwidth parameter for mean shift algorithm is set to $b_{init}$ and the number of clusters is observed. Then at each iteration, the bandwidth is increased by $b_{step}$.  For all our images $b_{init}=N_c$ and $b_{step}=b_{init}/2$. The iterations are stopped whenever the number of clusters $k$ drop below $N_c$. The resulting $k$ clusters are used to define $k$ binary images. The resulting bandwidth is taken as the image dependent bandwidth estimate $b^*$. \\

\noindent {\bf{Refinement.}}  The computation of  initial clusters as outlined above is performed in the color space. Hence, spatial proximity of the pixels are not taken into account. In the next stage,  a sequential combination of median filtering in the pixel space and mean shift clustering in the color space are applied iteratively for the fixed  bandwidth $b^*$.  
The median filtering is realized as follows:  If a pixel of class $c_i$ is surrounded by a pixel of class $c_j$, it is assigned to class $c_j$, and 
$L*a*b*$ channels of that pixel to the cluster center of the cluster $c_j$.  This sequential application of median filtering followed by mean shift with fixed bandwith $b^*$ is performed only few  times. For all of our images, five iterations seemed sufficient. 
More iterations may cause the components of different colors to join.  

Final clusters in the pixel space may have small holes. These holes may result either from an insufficient application of the iterative and sequential filtering step outlined above or simply from a small feature such as an eye of a bird. 
To remove holes, all background (foreground) connected components with radius smaller than a given threshold $R$ are converted to  foreground (background) pixels.  Performing elimination based on component radius rather than component area is more reliable, because it might be the case that all (or some) of the components join giving large areas causing the necessary but separate components to be eliminated. 

\begin{figure}[ht]
\centering
  \begin{tabular}{l}
  \includegraphics[width=1\linewidth,trim={0 0 0 0},clip]{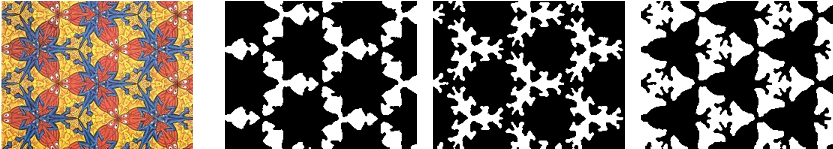}
\end{tabular}
\caption{A sample result for image processing module. The input ornament image (left) and three masks.}
\label{fig:seg_example}
\end{figure}

A sample result of the image processing module is depicted in Fig.\ref{fig:seg_example}.\\
%

\subsection{Local connectivity analysis I: from masks to connections} \label{ssec:centroids}
\noindent The process of local connectivity analysis starts with consistent keypoint detection on connected components.  An obvious means is to detect centroid of each  foreground component. However, this will be very sensitive to the output of the image processing module. Separate repeating motifs may touch one another, for example.  Furthermore, this joining of the motifs may be inconsistent throughout the ornament plane. What is important for the purpose of further analysis is that the detected keypoints are consistent throughout the pattern. It is, however, not critical whether they really coincide with the true motif centers. Hence, we call these keypoints as {\sl nodes}. 

\begin{figure}[ht]
\centering
  \begin{tabular}{l}

  \includegraphics[width=1\linewidth,trim={0 0 0 0},clip]{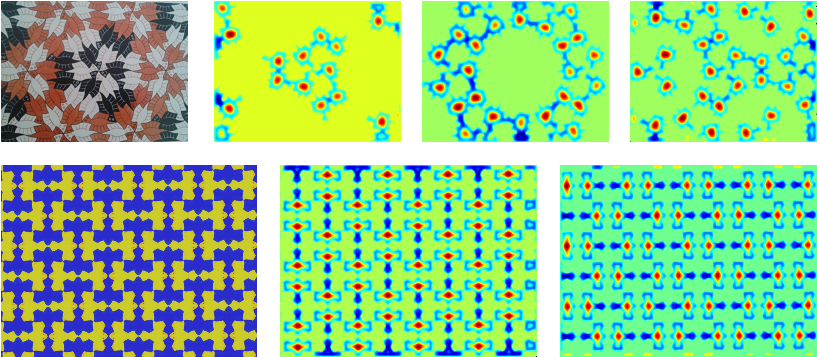}

\end{tabular}
\caption{Sample label images. Observe that the motif centers obtain the highest values.}
\label{fig:labeling}
\end{figure}

Towards robustly locating nodes,  a continuous labelling on each binary image, namely mask, is performed, yielding a continuous image 
$label^*$, whose values are in $(-1,1)$. The labelling stage is detailed at the end of this subsection as a separate paragraph.  The nodes are calculated as the centroids of the positive valued connected components of the  $label^*$ image. Some sample label images are depicted in Fig.\ref{fig:labeling}.  A node is  merely a robustly computed keypoint. 

Once the nodes are detected, a graph called connectivity graph is constructed via an  iterative extraction of local node relations. 
In the first iteration, the minimal pairwise distance is found, and then all connections with similar distances are extracted (using a fixed tolerance $tol$). In the next iteration, excluding the extracted connections,  next minimal distance is computed to extract new connections. After  $n$ iterations, connections of various sizes are obtained. Note that  the connections with large distances are not much meaningful; at best they  provide redundant information. Hence, it is better to keep $n$ small. Neither the choice of the parameter $n$ nor the tolerance $tol$ is critical, because we later re-classify the connections using  the mean shift algorithm. 

Given a set of connections stored at the connectivity graph, the connection length as the feature is fed to the mean shift. A small and fixed  bandwidth, $bc^*$ is used. The number of connection groups as discovered by the mean shift clustering could be different than the initial $n$. 

\begin{figure*}[ht]
\centering
  \begin{tabular}{cc}

  \includegraphics[height=0.27\linewidth,trim={0 0 0 0},clip]{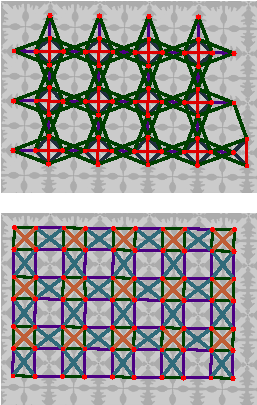}&
   \includegraphics[height=0.27\linewidth,trim={0 0 0 0},clip]{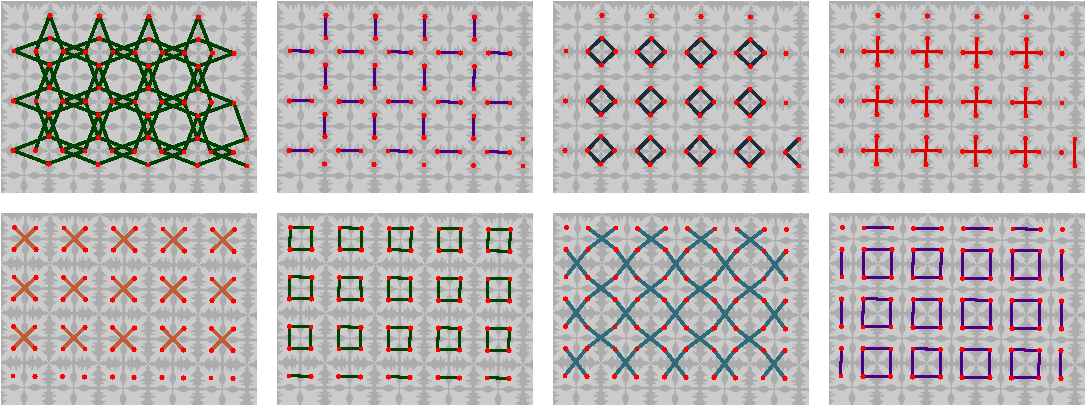}\\
a & b
  \end{tabular}
\caption{Connectivity graph. (a) Extracted connections for each mask of the ornament, (b) connection groups.}
\label{fig:graph}
\end{figure*}

In Fig.\ref{fig:graph}, a sample connectivity graph and individual connection groups, discovered by clustering using the connection length, is depicted. \\
%

\noindent {\bf{Continuous labelling on the binary masks.}} For each mask of each ornament,  each pixel $p$ on each foreground connected component is assigned an initial label, $label(0,p)$, reflecting whether or not the probability of the pixel belonging to a centroid is higher than that of the pixel  belonging to the component boundary. For this purpose, distance from each foreground pixel to the nearest  background pixel is computed, and then those which have bigger distance than the half of the maximum of all distances are assigned a positive constant $a$ where the  others $-a$. 

After this label initialization,  a relaxation is  performed. In the relaxation step, the value is increased (decreased) depending on whether the current label at a pixel $p$ is less (more)  than the average of neighboring labels.  Formula-wise, relaxation is expressed as follows:
 
\[ label(i+1, p) = label (i,p) + r \ast  relaxation \]
where $r$ is the relaxation constant which we take as positive, and the $relaxation$ is
\[
relaxation=     \mbox{Avg} \left( label(i,p) \right)  -  label(i,p) + label(0,p) 
\] 
where the $\mbox{Avg} \left( label(i,p) \right) $ is computed as the average of $4\mbox{-neighboors}$. 
At convergence or after sufficient number iterations, whichever comes first,   the continuous labelling of binary masks is achieved as 
\[ label^*(p) \leftarrow label(K,p) \]
where $K$ is last performed iteration.
\subsection{Local connectivity analysis II: from connections  to symmetry elements }
\noindent The connection groups extracted as described above are further analyzed in order to detect node relations. The way the nodes are related gives a hint on various symmetry elements. Recall that the connections with small sizes are favored, since larger connections repeat the same node relations at a larger scale. Thus, the analyses begin with the connection group of the smallest size and continue in ascending order. The analyses are done as following.

Each individual connection group is divided into $connected$ $graphs$. Afterwards, each connected graph is further analyzed independent of the other connected graphs in the connection group. Given a connected graph, the following decisions are made.

\begin{figure}[ht]
\centering
  \begin{tabular}{c}

    \includegraphics[width=1\linewidth,trim={0 0 0 0},clip]{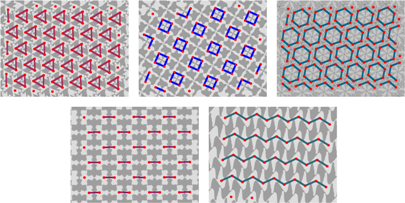}

  \end{tabular}
\caption{Examplar connections. Top row illustrates cyclic graphs, while acyclic ones are shown at the bottom row. Equilateral triangles indicate three-fold rotations, squares indicate four-fold rotations, and regular hexagons indicate six-fold rotations. Two-point connections represent two-fold rotations and zigzag structures indicate glide reflections.}
\label{fig:graph_pex}
\end{figure}

\begin{itemize}
\item[1] If a graph is a $cycle$ $graph$, $i.e.,$ the nodes are connected in a closed chain, then the probability of the graph being either equilateral triangle, square or regular hexagon is computed (Fig.\ref{fig:graph_pex}(top row)). The probability is the product of ratios of polygon edges. Thus, if a graph is constructed from the connection of three (four, six) nodes, and the probability computed as a product of its edge ratios is higher than by chance ( $i.e.$, 0.5), the center of the triangle (square, hexagon) is taken as a center of three-fold (four-fold, six-fold) rotation.

\item[2] If the graph is acyclic then the nodes might be related either by two-fold rotation (a graph containing only two nodes) or glide reflection (a graph containing nodes connected in a zigzag form)  (Fig.\ref{fig:graph_pex}(bottom row)). 
\begin{itemize}
\item[2a] If the graph contains only two nodes, their center is taken as the center of two-fold rotation. Since two-node connections might occur by accident, the two-fold centers are accepted only if the number of nodes involved in such connection in a connection group is more than $60 \%$ of entire node number in a mask.

\item[2b]If the graph contains more than three nodes, the polynomial of order one is fit to the given nodes. For a graph of zigzag structure the line should pass through the centers of the edges between two adjacent nodes. In this case, the distances between the centers of the edges of two adjacent nodes and a line are computed. The probability of the graph being of zigzag structure is computed by taking the product of the above distance ratios. If the probability is more than 0.5, then there is a glide reflection axis passing through edge centers. On the contrary, if the nodes themselves (not the edge centers) lie on the fit polynomial, then this is a line, representing translational symmetry. \\
\end{itemize}

\end{itemize}

\begin{figure}[ht]
\centering
  \begin{tabular}{m{3cm}m{3cm}}

\multicolumn{2}{c} {\includegraphics[width=0.9\linewidth]{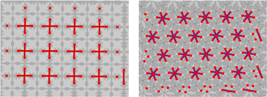} }\\
           \hspace{1.5cm}(a)& \hspace{1.6cm}(b)
           \\
\multicolumn{2}{c} {\includegraphics[width=0.9\linewidth]{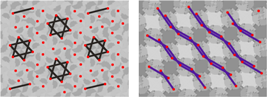} }\\
           \hspace{1.5cm}(c)& \hspace{1.6cm}(d)\\
   \end{tabular}
   \caption{Implicit node relations. (a) Double two-fold rotation centers, (b) triple two-fold rotation centers, (c) double three-fold rotation centers, (d) double glide reflections. }
   \label{fig:implicit_conn1}
   \end{figure}

\begin{figure}[ht]
\centering
  \begin{tabular}{c}

  \includegraphics[width=0.9\linewidth,trim={0 0 0 0},clip]{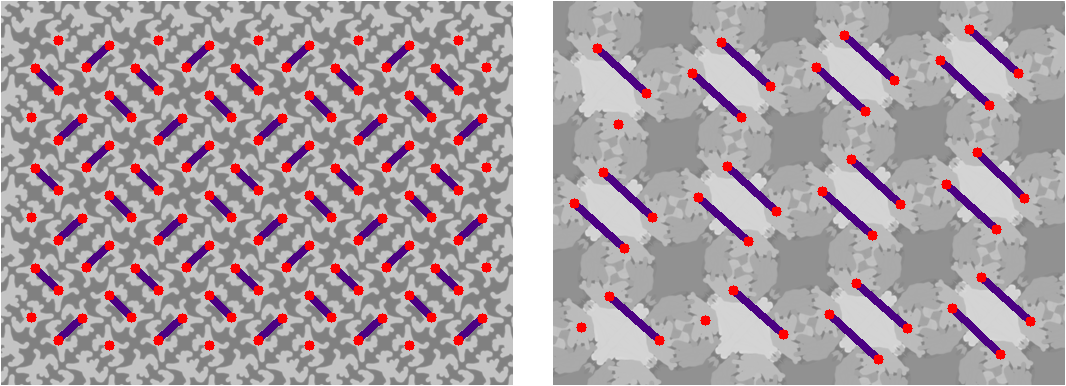}
   \end{tabular}
   \caption{Samples for paired two-fold connections. The actual two-fold rotational centers lie at the centers of paired lines.}
   \label{fig:implicit_conn2}
   \end{figure}

\noindent {\bf{Detecting implicit node relations.}} After defining symmetry centers, elimination of repeating symmetry centers is performed. For example, if from early connection group a point on a mask has been marked as a center of four-fold rotation, and the same point is again marked as four-fold rotation center from subsequent connection group, the later one is discarded. This means that a symmetry center of one type is obtained only from one connection group. However, the symmetry center of one type may coincide with the symmetry centers of other types. For example, the symmetry center marked as four-fold rotation center from some connection group, may also be the symmetry center of two-fold rotation derived from some other connection group. A point in a mask may represent multiple symmetry centers of the same type, if and only if they are derived from the same connection group. Thus, two three-fold rotation centers (double triangle) at the same point might occur only when those are detected from the same connection group. Such repetitions are used to detect implicit node relations. If there is a double triangle at a point, this indicates that there is a six-fold rotation around that point. Similarly, if there is double two-fold (triple two-fold) rotation center at the point this is an indicator of four-fold (six-fold) rotation around that point if the angle between two-point lines is $90 \degree$ ($60 \degree$). Double glide reflection axes represent reflection symmetry which is perpendicular to the glide reflection axis. Samples of such connections are shown in Fig.\ref{fig:implicit_conn1}.\\
Another case is paired two-point connections (see  Fig.\ref{fig:implicit_conn2}). They too indicate reflection symmetry. Actual two-fold centers lie at the centers of those paired lines. Moreover, they are the indicators of reflection axis passing through two-fold rotation centers. The algorithm detects them by computing the minimal distance between two-fold rotation centers and connecting nodes that are in this minimal distance from each other. If a graph of maximal degree four is obtained then these are actual two-fold rotation centers. However, if two-point connections occur then these are paired two-fold rotation centers and are handled accordingly.\\

\begin{figure}[ht]
\centering
  \begin{tabular}{m{3cm}m{3cm}}

\multicolumn{2}{c} {\includegraphics[width=1\linewidth]{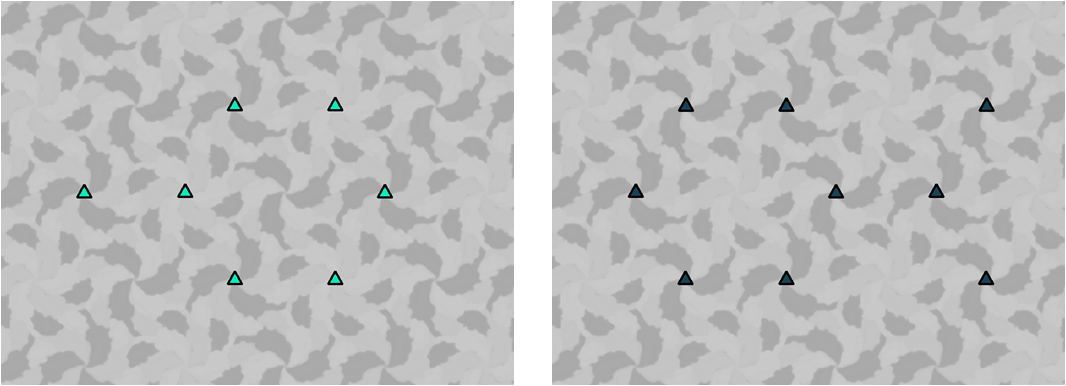} }\\
           \hspace{1.7cm}(a)& \hspace{2.2cm}(b)
           \\

\multicolumn{2}{c} {\includegraphics[width=1\linewidth]{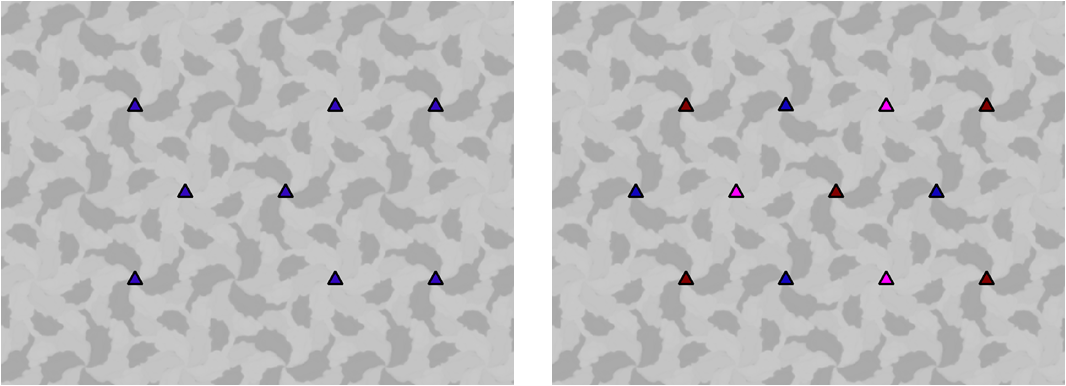} }\\
           \hspace{1.7cm}(c)& \hspace{2.2cm}(d)\\
   \end{tabular}
   \caption{(a-c) An example of symmetries detected for three masks of an ornament. For each mask one class of three-fold rotation centers are detected. (d) when the symmetries of all three masks are collected,  three classes of three-fold rotation centers are obtained.}
   \label{fig:sym}
   \end{figure}

\noindent{\bf{Refinement.}} Notice that up to now the symmetries for each mask are detected. After these steps, the algorithm yields a mask structure which contains fields for various symmetry centers and their classes. For example, if a mask contains three-fold rotation centers, the structure contains their centers and classes. The class of symmetry element is the number of connection group from which it was extracted. Thus, all three-fold rotational centers extracted from the same connection group are of the same class. The next step is to collect all symmetries detected on individual masks. This is a necessary step, since each mask may contain only one class of some symmetry type, while a whole pattern contains more classes of that symmetry type. Initially, all symmetries of the same type are collected without considering their classes. Say, the three-fold rotation centers extracted from all masks are marked on an ornament. Then for each center, the algorithm defines which classes fall into that particular center. Thus, if three-fold rotation centers of the first mask are of class 2, 5 for mask two and 3 for mask three, the centers contain classes 2 and 5 (25), 2 and 3 (23), and, 5 and 3 (53). These numbers define the classes of the symmetry centers and they are classified accordingly into three groups. This example is illustrated in Fig.\ref{fig:sym}. It might be the case that the symmetry center is detected only for one mask, and that center has not been detected for another masks. Then this center forms a group on its own, leading to fourth symmetry class. To eliminate such groups, the algorithm counts the number of centers of each class, sorts them in ascending order, computes the minimal distance between the nodes in one class, and beginning with the largest class propagates its group to the nodes that are of similar distance. This is done iteratively, until no changes occur.\\
Once all the symmetry types are combined and their classes are determined the maximal order of rotation is defined. If no rotations are observed then the maximal order of rotation is taken as $1$ and it is checked whether the pattern contains glide reflections. If no glide reflection observed then the tile is of $p1$ group, {\sl i.e.},  contains only translational symmetry. 
After defining the maximal order of rotation, further elimination is done using the symmetry group information. Thus, if the maximal order of rotation is four or six then there can be only one class of two-fold rotation centers. If, erroneously, two classes were detected, the class intersecting with the unit cell edges is left, the other type is eliminated. If the maximal order of rotation is six and there are more than one classes of three-fold rotations, then the one intersecting with the unit cell centers is left, and the others eliminated.
\subsection{Final symmetry group detection}
\begin{figure*}[!htb]
\centering
\includegraphics[width=0.9\linewidth, trim={0 0 0 0}, clip]{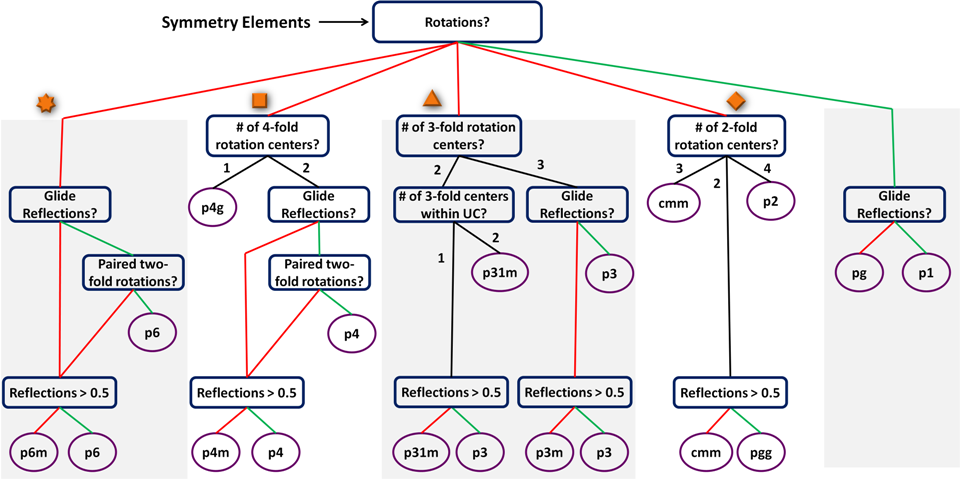}\\
\caption{Proposed decision tree. }
\label{fig:decision_tree}
\end{figure*}

\noindent Individual cues in the form of symmetry elements as described in the previous subsection are integrated to yield the final symmetry group decision via a decision tree which we propose for the reduced set of $13$ groups, The tree is  shown in Fig.~\ref{fig:decision_tree}. 
Comparing our tree to the classical decision tree in Fig.~\ref{fig:decision_tree_classic}, observe that mirror reflection checks are postponed to be performed at the last stage and their number is significantly  reduced.  Furthermore, with the exception of two-fold rotations case, a mirror reflection check is  performed only when other cues indirectly imply it.
For example, if the maximal order of rotation is six and glide reflections or paired two-fold rotations are detected, then a mirror reflection check is in order. 
If the probability of mirror reflection is calculated to be more than $0.5$ then the ornament  is classified as belonging to  $p6m$ group and else to  $p6$ group. 
The reason has been explained before in the Introduction section: An  erroneously detected mirror reflection is less desirable  than a missed one.  In case of a missed mirror reflection, the fundamental domain will just be twice as big as it really should be, the half of it being the mirror of the other half; hence, the whole pattern can be correctly generated. 

There is one  exception: The case when  two two-fold classes are detected. In this case, there are two possibilities.
The first possibility is that the ornament is of group $pgg$. The second possibility is that the ornament is a $cmm$ pattern formed by all mirror symmetric protiles. 
In the latter case,   the third two-fold center for $cmm$ is missed because  it  happens to be  on the glide reflection axis and cannot be captured from the connection graphs. 
Hence, a $cmm$ tile is classified as $pgg$ due to a missed third two-fold center. 
This is not tolerable because unlike $p6m$ versus $p6$ or $p4m$ versus $p4$, the mirror reflection is not the only distinction between the two groups of $cmm$ and $pgg$, hence, the fundamental domain is not correct up to a mirror.
For this reason, in the case of two two-fold centers, 
  the mirror reflection check is performed  even if it is not implied via indirect cues. \\

\noindent{\bf{Mirror Reflection.}} Mirror reflection checks are performed on a single unit cell.  Note that, in case of rotational groups of three, four or six fold orders,  the unit cells of the regular and reflectional groups are identical, the difference being in the fundamental domain.  In the case of two-fold rotations, the unit cell of the $pgg$ group is employed.  At any rate, given a unit cell, all the objects lying in it are extracted using the masks of the ornament pattern. Then the unit cell is divided into two along the expected reflection axis (based on group information).   For each object lying on both sides, the area ($A$), the perimeter ($P$), the distance of the object center to the reflection axis ($dRef$), the distance of the farthest point on the object to the reflection axis (fdRef) and the point on the reflection axis which is the closest to the object center ($pRef$) are computed. 
A pair of objects, one from the first part of the unit cell ($obj1$) and the other from the second  ($obj2$),  is picked. The probability of the first object being a reflection symmetry of the second object is estimated via the product of the corresponding feature ratios:
\begin{eqnarray}
 prob=&\frac{A_{obj1}}{A_{obj2}} * \frac{P_{obj1}}{P_{obj2}} *\frac{dRef_{obj1}}{dRef_{obj2}}*\frac{fdRef_{obj1}}{fdRef_{obj2}} \nonumber \\
&* \exp(\frac{-d(pRef_{obj1},pRef_{obj2})}{\sigma}) \nonumber
\end{eqnarray}
In all experiments, $\sigma=10$. 
For each object in the first part, the above probability is computed for all its pairs. The highest score is picked. It  indicates with what probability the  object has a mirror object  on the other side. Then, the mean of all the highest scores in the first part is taken to indicate with what probability the unit cell is symmetrical along the given reflection axis.  Note that there might be more than one reflection axes for a given symmetry group. The final probability is the mean value computed for all reflection axes.\\

\noindent {\bf{Unit Cell.}} Recall that, in all rotational groups, the corners of the units cells are the points of  maximal order of rotation. With the exception of  $p4g$ and $pgg$, the unit cell is  readily constructed by connecting the nearest maximal rotation centers of the same class.  For the two groups, $p4g$ and $pgg$, however, the maximal rotational  symmetry center appearing on the unit cell corners also appears on the unit cell center. Therefore, for the latter two groups, a point $pt$ of certain class is chosen, then four closest points to $pt$ of similar class are selected. Among these four points $pt1$ and $pt4$ are selected such that the length of $(pt,pt1)$ equals to the length of $(pt,pt4)$ and are in opposing directions. Then the rest two points $pt3$ and $pt2$ should also be at equal distance to the point $pt$ and be in opposing directions. If all points are exhausted, the algorithm reports a failure.\\

If there are no rotations but glide reflections, then the lattice nodes are centers of two-point connections of the zigzag structure with similar directions. To detect a unit cell, a point ($pt1$) of particular class is chosen as a first node for unit cell. Then two closest points ($pt2$ and $pt3$) to $pt1$ are chosen so that the angle between lines $(pt1,pt2)$ and $(pt1,pt3)$ is less than or equal to $90 \degree$ and bigger than or equal to $60 \degree$. The last point $pt4$ is chosen so that the angle between lines $(pt4,pt2)$ and $(pt4,pt3)$ is equal to the angle between lines $(pt1,pt2)$ and $(pt1,pt3)$, and the length of lines $(pt4,pt2)$ and $(pt4,pt3)$ are equal to lines $(pt1,pt3)$ and $(pt1,pt2)$, respectively. Recall that all lattice nodes are of the same class.  If no such points detected, then another point is chosen as $pt1$ and the same operations performed according to this point. If all points have been exhausted, then the algorithm reports a failure.
If an ornament pattern has neither rotations nor glide reflections then the unit cell nodes are merely the motif centers. \\

\noindent {\bf{Fundamental Domain.}} If the unit cell is successfully constructed, then the fundamental domain extraction is straightforward,
 recall Fig.~\ref{fig:cell_struct}. For example, the fundamental domain for a $p6$ ornament is $1/6^{th}$ of its unit cell whereas it is  $1/12^{th}$  for a  $p6m$ ornament.
If, however, the algorithm reports a failure during unit cell construction, then algorithm returns to the stage before the attempt to construct a unit cell  to make use of all the previously collected information on individual symmetries. (Note that failure during unit cell construction may arise due to lack of sufficient translational repetition.) Let us explain the fundamental domain extraction step in case of failure via an example. 

Suppose only two centers of six-fold rotation ($pt1$, $pt2$) are found, hence, the unit cell cannot be constructed.  
Assuming that this two centers are both nodes of the same unit cell the distance $rd$ between these two points is computed. Since it is known that the unit cell for ornaments with six-fold rotations is composed of two equilateral triangles, there should be a three-fold center ($pt3$) which is in a $\sqrt{3}rd$ distance from the both points.
 In order to detect $pt3$,  two circles of radius $\sqrt{3}rd$ centering at $pt1$ and $pt2$ are drawn. There will be two points at which this two circles intersect. Any one of them can be selected as $pt3$. In this manner, three points indicating nodes of a fundamental domain are computed. For the cases when more than two six-fold rotations exist, two closest nodes are selected as $pt1$ and $pt2$.  If, on the other hand, only one six-fold rotation center ($pt1$) is found, then three-fold centers are searched. Then the fundamental domain is extracted using the relation of six-fold center and three-fold center ($pt2$) that is closest to the six-fold center. A point $pt3$ is selected so that the length ($pt3,pt2$) is equal to the length of ($pt1,pt2$) and the angle between these two lines is $120 \degree$.  If no three-fold centers are found then two-fold centers are searched. Then the two-fold center ($p$) closest to the six-fold center is used to extract the rest of the fundamental domain. In this case $pt2$ is a point which is equal to $d(pt1, p)$ and in opposing direction. This new point is actually a six-fold center. Then $pt3$ is found using the first case where two six-fold centers are known.  If no other symmetries detected except for one six-fold center then algorithm fails to detect the fundamental domain. Note that it might be the case that two six-fold centers are found in two different locations so that in between six-fold centers are missed by the algorithm. In such cases the first case gives large fundamental domain. If more than one type of symmetries detected so that more than one cases described above hold, then the fundamental domain is computed for all of the cases. The final fundamental domain is the one with the smallest region. 
 
This  approach is extended to other symmetry groups using their own properties.

\section{Experiments}
\label{sec:results}
\subsection{Data}
\noindent We formed a labelled ornament data set produced under different imaging conditions by a variety of ornament artists including the authors {using iOrnament tool \cite{iornament}.}  This set is enriched with 14 representative ornament fragments that are equivalent to $p1$, $p2$,$pg$, $pgg$, $p4$, $p4g$, $p3$, $p3m1$, $p31m$ if  color permutations are ignored. The total number of ornaments in the set add up to 100. In forming the data set, we paid attention to cover a variety of styles in terms of brush, color, tone, and motif choice. We further paid attention that half of the ornaments are mimicking Escher's style with asymmetric interlocking forms and the set contains enough representative elements in each of the $13$ groups.

\subsection{Parameters}
\noindent For all $100$ ornaments, identical parameter values are used as detailed below. \\

\noindent {\bf{Gamma correction.}}  $\gamma$ is set to $0.5$.\\

\noindent {\bf{Clustering.}} $N_c$ is set to  $10$. For the refinement stage,  $R$ is set to $R=0.3*maxR$, where $maxR$ is the maximum radius of the connected components.  \\

\noindent {\bf{Connection extraction.}} The connectivity graph is constructed using  $n=5$ iterations with $tol=5$. The bandwidth fed to mean shift for connection group clustering is set to $bc^*=2$.

\subsection{Results}
\begin{figure*}[!htb]
\centering
  \begin{tabular}{m{2.5cm}m{2.5cm}m{2.5cm}m{2.5cm}m{2.5cm}m{2.5cm}}

\multicolumn{6}{c} {\includegraphics[width=1\linewidth]{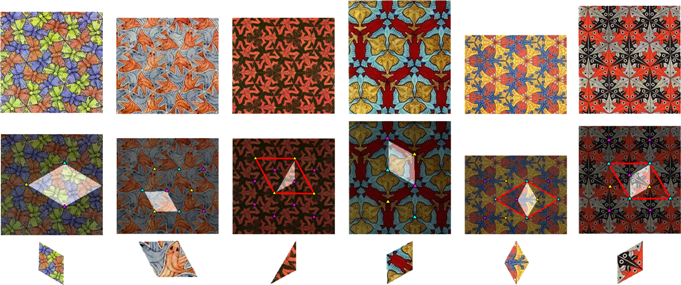} }\\
             \hspace{1.1cm} (a) $p3$& \hspace{1cm}(b) $p3$& \hspace{0.8cm}(c) $p31m$& \hspace{0.5cm}(d) $p3 \tcr{m}$&\hspace{0.5cm} (e) $p3 \tcr{m}$&  \hspace{0.5cm}(f) $p3 \tcr{m}$\\
\\
\multicolumn{6}{c} {\includegraphics[width=1\linewidth]{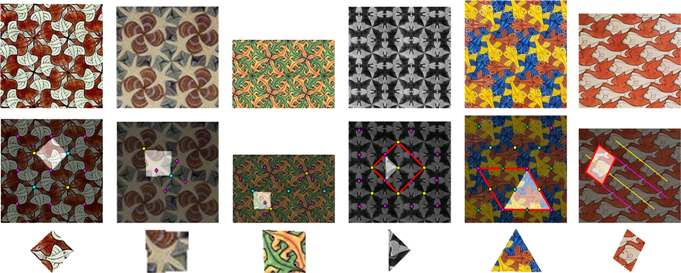} }\\
             \hspace{1.1cm} (g) $p4$&\hspace{0.8cm} (h) $p4$& \hspace{0.8cm}(i) $p4$&\hspace{0.5cm} (j) $p4g$& \hspace{0.5cm}(k) $p2$&\hspace{0.5cm} (l) $p1$\\ \\\\
              
\multicolumn{6}{c} {\includegraphics[width=1\linewidth]{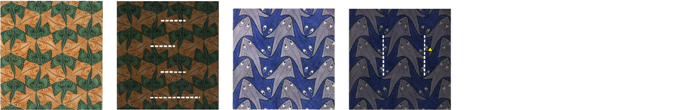} }\\
\hspace{1.1cm}(m)& &\hspace{0.8cm}(n)

%
 \end{tabular}
\caption{Symmetries detected for ornaments painted by Escher. Observe that although most of the images are small  and have few repetitions of the symmetries, we are able to find the fundamental domains. Having insufficient repetition seems to pose a problem only to two glide group ornaments shown in the last row.  The letter $m$ in the group names under the ornaments are shown in red if the mirror reflection is missed. In those cases, the fundamental domains are double the sizes they should be. }
\label{fig:results_escher}
\end{figure*}

\noindent We first present our result on $14$ representative ornaments taken from Escher's collection  (Fig.~\ref{fig:results_escher}).  For the first $12$ out of $14$ ornaments, depicted in Fig.~\ref{fig:results_escher} (a) through (l),  our method works successfully. In each group, (a) through (l),  the first rows show the original input. The second rows show the detected symmetries, the unit cells and the fundamental domains superimposed on the  input. The third rows show the fundamental domains cut out automatically from the original patterns.  
Unit cells (in the forms as previously described in Fig.~\ref{fig:cell_struct}) are shown by red quadrilaterals. 

{For illustration purposes, the region belonging to the fundamental domain is made lighter while the rest of the pattern is made slightly darker. 
In case an ornament image is too small to fit a whole unit cell in it, only the fundamental domain is shown. Half of the cases ($6$ out of $12$) are like this. }

Notice that in some cases, (d)-(f), the letter $m$ of the group names are  in red color. This is to indicate that our method in these three cases missed the mirror reflection.  
For none of these three examples, mirror reflection checks at the last stage  are performed because indirect clues did not imply mirror reflection. 
As a result, the fundamental domains are twice as big; though, the unit cells are correct. 
{Recall that in general mirror symmetry can not cause any problems, since the centers of the maximal order of rotations do not reside on the centers of the respective protiles.}

{For the famous  mariposas pattern (Fig.~\ref{fig:results_escher} (a)) only three three-fold rotation centers all from different classes are detected. 
For the second $p4$ tile (Fig.~\ref{fig:results_escher} (h)) only two four-fold rotation centers both from different classes are detected. Yet, for both patterns we obtain enough information to identify their symmetry groups and detect fundamental domains. The results in Fig.~\ref{fig:results_escher} (h)-(i) show two-fold rotation centers that are detected in wrong places. Nevertheless, since the symmetry group depends on the maximal rotation order, incorrect two-fold centers do not influence final decision. }

For the last two ornaments, respectively in $pgg$ and $pg$ groups, the collected cues are not sufficient. In each case, only one of the glide axes are detected. Hence, these two  cases are inconclusive. The reason for the failure is that it is harder to detect glides in the absence of sufficient repetition. As we demonstrate in later examples, detecting $pg$ or $pgg$ is possible when there are slightly more samples.
 
\begin{figure*}[!htb]
\centering
  \begin{tabular}{m{2.7cm}m{2.7cm}m{2.7cm}m{2.7cm}m{2.7cm}}

 \multicolumn{5}{c} {\includegraphics[width=1\linewidth]{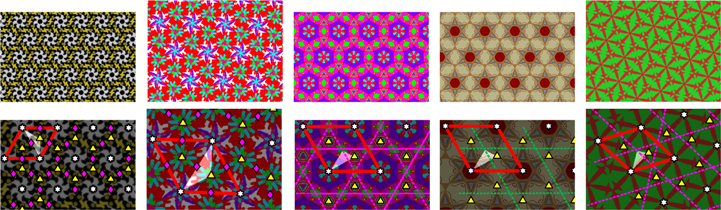} }\\
        \hspace{1.5cm}$p6$& \hspace{1.5cm}$p6$& \hspace{2cm}$p6m$& \hspace{2cm}$p6m$& \hspace{2cm}$p6m$\\
\\

 \multicolumn{5}{c} {\includegraphics[width=1\linewidth]{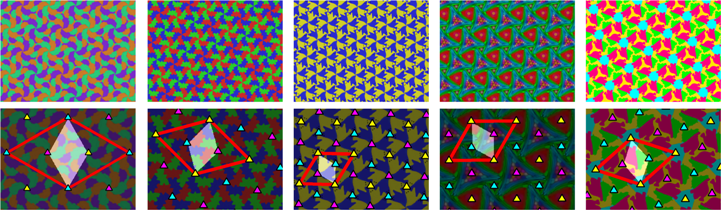} }\\
          \hspace{1.5cm}$p3$ &  \hspace{1.5cm}$p3$ &  \hspace{2cm}$p3$ &  \hspace{2cm}$p3$ &  \hspace{2cm}$p3$\\
\\

 \multicolumn{5}{c} {\includegraphics[width=1\linewidth]{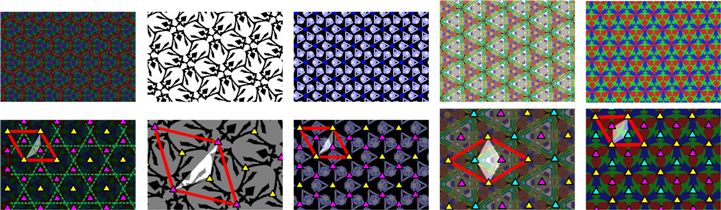} }\\
        \hspace{1cm} $p31m$ &  \hspace{1.5cm}$p31m$ &  \hspace{1.5cm}$p31m$ &  \hspace{2cm}$p3 \tcr{m}$ &  \hspace{2cm}$p3 \tcr{m}$\\
         
                 \end{tabular}
\caption{Results for the ornaments with six- and three-fold rotations. }
\label{fig:results_triangle}
\end{figure*}
\begin{figure*}[!htb]
\centering
  \begin{tabular}{m{2.7cm}m{2.7cm}m{2.7cm}m{2.7cm}m{2.7cm}}

\multicolumn{5}{c} {\includegraphics[width=1\linewidth]{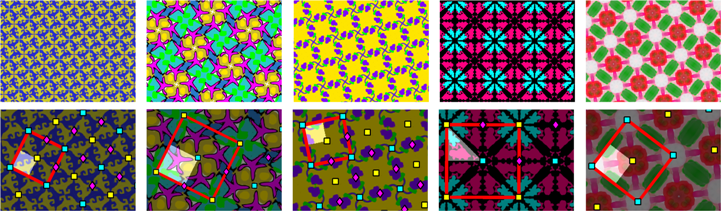} }\\
          \hspace{1.2cm}$p4$ &  \hspace{1.5cm}$p4$ &  \hspace{2cm}$p4$ &  \hspace{2cm}$p4m$ &  \hspace{2cm}$p4 \tcr{m}$\\
\\

\multicolumn{5}{c} {\includegraphics[width=1\linewidth]{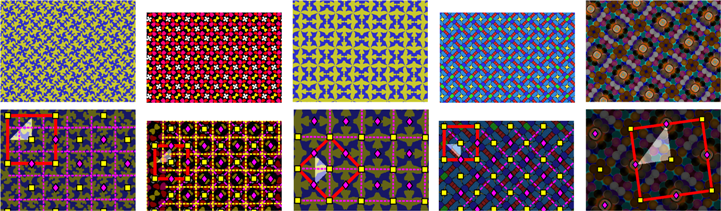} }\\
            \hspace{1.2cm}$p4g$ &  \hspace{1.5cm}$p4g$ & \hspace{1.5cm} $p4g$ &  \hspace{2cm}$p4g$ &  \hspace{2cm}$p4g$\\
\end{tabular}
\caption{Symmetries detected for ornaments  with four-fold rotations.}
\label{fig:results_4fold}
\end{figure*}

{The rest of the results for the remaining $85$ tiles are organized as follows. (Note that the result for one $p6$ ornament had been shown previously in the Introduction.) To save space, $45$ samples are placed in the Appendix. They are split into three figures each showing the results for $15$ patterns.
To be illustrated in this section, $25$ illustrative samples  from $8$ groups of higher order rotations are selected. The remaining $15$ are  selected from the remaining $5$ groups. The patterns with higher order rotations are further organized into two groups. The first one contains  the groups with triangular lattice structures ($p6m,\,p6,\,p31m,\,p3m,\,p3$); for these groups the unit cell consists of two equilateral triangles.
The other contains the groups with square lattice ($p4m,\,p4g,\,p4$). The results are respectively shown in Figs.~\ref{fig:results_triangle} and \ref{fig:results_4fold}. Original ornaments are shown in their full sizes whereas  the ones depicting the  results are cropped in order to make the symmetries, unit cells and fundamental domains visible.}

 In all samples of the higher order rotation groups, the symmetry groups are correctly identified, except  up to mirror reflection in some samples.   Because a $p31m$ ornament has two three-fold rotation centers,  it is automatically classified correctly without a need for mirror reflection check.
In our $p6m$ examples, the mirror reflections were implied as a result of detected glides, so  the flow in the decision tree proceeded to mirror reflection check; hence, the fundamental domains are correctly identified.  The fundamental domain for a  $p6m$ ornament  is half the size of a $p6$ one,  and that of a $p3m$ ornament is 
half the size of a $p3$ one. This is because the other halves can be obtained by mirror reflection. In our $p3m$ examples, the fundamental domains are double the size they should be. This is because the samples are classified as belonging to $p3$ as a consequence of missed mirror reflections. 
Nevertheless, using the generation the rules for the $p3$ group, instead of for the $p3m$, the original tiles can still be recreated.
 
\begin{figure*}[!htb]
\centering
  \begin{tabular}{m{2cm}m{2cm}m{2cm}m{2cm}m{2cm}m{2cm}m{2cm}}

                        \multicolumn{7}{l} {\includegraphics[width=0.9\linewidth]{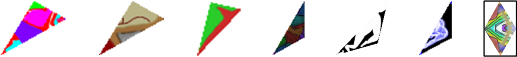} }\\
        \hspace{0cm} $p6m$&\hspace{0.5cm}$p6m$&\hspace{0.5cm}$p6m$&\hspace{0.5cm}$p31m$\hspace{0cm}&$p31m$&\hspace{0cm}$p31m$&\hspace{-0.5cm}$p3 \tcr{m}$\\
\\              

 \multicolumn{7}{l} {\includegraphics[width=0.9\linewidth]{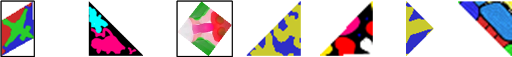} }\\
                         \hspace{0.1cm} $p3 \tcr{m}$& \hspace{0.5cm}$p4m$& \hspace{0.6cm}$p4 \tcr{m}$&  \hspace{0.5cm}$p4g$ & $p4g$ & $p4g$ & \hspace{-0.5cm} $p4g$ \\
             \\

 \multicolumn{7}{l} {\includegraphics[width=0.9\linewidth]{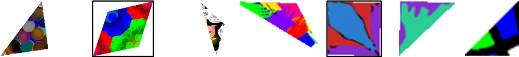} }\\
                             \hspace{0cm} $p4g$& \hspace{0.5cm}$p3 \tcr{m}$& \hspace{0.5cm}$p31m$& \hspace{0.5cm}$p31m$& \hspace{0cm}$p4 \tcr{m}$& \hspace{0cm}$p4m$& \hspace{-0.5cm}$p4g$\\
         
                 \end{tabular}
\caption{Fundamental domains for ornaments  with mirror reflections.}
\label{fig:results_mirrors}
\end{figure*}
\begin{figure*}[!htb]
\centering
  \begin{tabular}{m{2.7cm}m{2.7cm}m{2.7cm}m{2.7cm}m{2.7cm}}

\multicolumn{5}{c} {\includegraphics[width=1\linewidth]{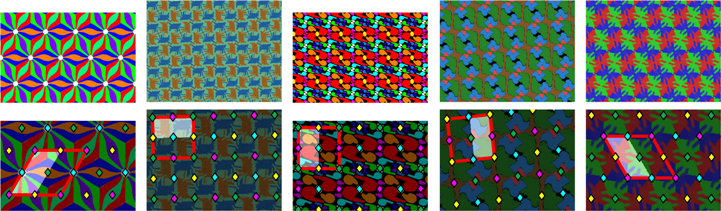} }\\
\hspace{1.2cm} $p2$&\hspace{1.5cm}$p2$&\hspace{1.7cm}$p2$&\hspace{2cm}$p2$&\hspace{2cm}$p2$\\
\\

\multicolumn{5}{c} {\includegraphics[width=1\linewidth]{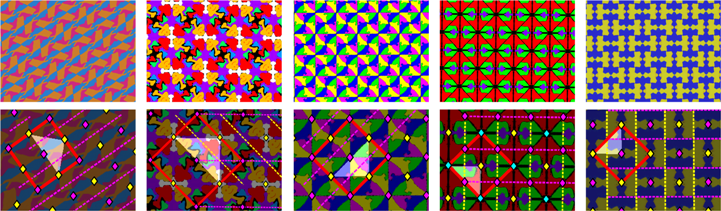} }\\
               \hspace{1.2cm}$pgg$&\hspace{1.5cm}$pgg$&\hspace{1.5cm}$pgg$&\hspace{1.7cm}$cmm$&\hspace{1.7cm}$cmm$\\ 
               \\

\multicolumn{5}{c} {\includegraphics[width=0.99\linewidth]{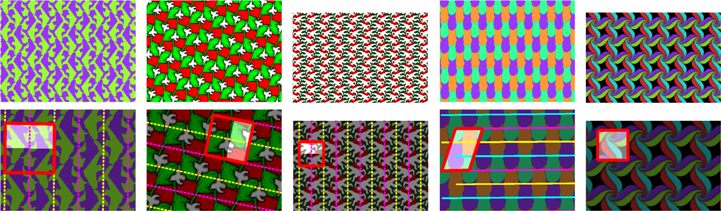} }\\
                        \hspace{1.2cm} $pg$&\hspace{1.5cm}$pg$&\hspace{1.7cm}$pg$&\hspace{2cm}$p1$&\hspace{2cm}$p1$
\\
 \end{tabular}
\caption{Sample results for ornaments in $p2,\,pgg,\,cmm,\,pg,\,p1$ groups.}
\label{fig:results_5g}
\end{figure*}

{As for the  two samples of $p4m$ (Fig.~\ref{fig:results_4fold}),  the first one successfully passed  the reflection double check (implied via  glides and then checked). But for the second sample, the  mirror reflection test is not performed because indirect clues did not indicate its existence. 
Hence, for this particular ornament of the $p4m$ group, the fundamental domain is double the size that it should be. Similar to the previously discussed  cases of $p3m$ ornaments, using the generation the rules for the $p4$ group, instead of for the $p4m$, the original $p4m$ pattern can still be recreated.} 

{As for the $p4g$ ornaments, their fundamental domains are accurately identified. In this group of ornaments, we  observed that glide reflections are easily detected, while for a human it is hard to perceive this type of symmetry.  For an ornament to be classified to $p4g$ group it is enough to detect one class of four-fold rotation centers. Hence, mirror reflection check is unnecessary. }

$25$ more results on tiles with higher order rotations are given in Appendix Figs.~\ref{fig:left} and \ref{fig:left2}. 
There are in total  $50$ ornaments out of $100$ with higher order rotations (excluding Escher ornaments), and $21$ of them  contain mirror reflections. The detected fundamental domains for these $21$ ornaments are given in Fig.~\ref{fig:results_mirrors}. The fundamental domains of the ornaments for which mirror reflections have not been detected are framed in black boxes. Observe that for $5$ ornaments, the mirror reflections are missed.  Hence, they are classified as belonging to the corresponding {\sl reflection-less} groups,  and their fundamental domains are double the sizes they  should be.

Fig.~\ref{fig:results_5g} shows results for the samples of the remaining five symmetry groups with lower order rotations: $p2$, $pgg$, $cmm$, $pg$, $p1$. The first five ornaments  belong to $p2$ group. This group of ornaments contain four distinct classes of two-fold symmetry centers. The second and the third groups respectively contain ornaments  with  two and three distinct classes of two-fold centers. Both of the groups ($pgg$ and $cmm$) have two distinct classes of glide reflection axes perpendicular to each other. The second sample of $cmm$ group (yellow and purple ornament) shows a case when the third class of the two-fold centers is not detected. 
This is because we employ two-point connections to indicate any binary connections, {\sl i.e.}, both the two-fold rotations and the glide fragments forming zigzag structures. 
In this particular $cmm$ example, our algorithm detects zigzag structures and defines them as glide reflection axes. During this process, however, our algorithm loses track of  the  two-fold rotations that are also indicated by the same two-point connections. 

In general, we observe that when two-fold rotation centers lie on the glide axes and all the protiles are mirror symmetric (as in the case of the present $cmm$ sample), our algorithm loses track of the one of the two-fold rotation centers. 
When one of the two-fold rotation centers is lost,  the symmetries of the $cmm$ pattern becomes similar to the symmetries of $pgg$ one except for the mirror reflection symmetry: while the $pgg$ group does not have mirror reflections, the $cmm$ group does. Thus, if exactly two two-fold rotation centers detected, we always need to perform mirror reflection check. For the present $cmm$ sample,  checking for the mirror reflection identifies the correct group.  We performed further tests with $6$ additional samples as presented in the 
Appendix Fig.~\ref{fig:left3}. With the help of  mirror reflection checks, our algorithm achieves $pgg$ and $cmm$ separation. 

Finally, the last row of Fig.~\ref{fig:results_5g} illustrates the results for ornaments without rotational symmetries. If two distinct classes of glides which are all parallel to each other are detected, then the ornament is classified as $pg$.  For the $p1$ group, the group of pure translation symmetry, the algorithm detects all grids or lines indicating translations.

\section{Summary and Conclusion}
\label{sec:conclusion}
\noindent We have presented a fully automated method to detect the symmetry group and extract fundamental domains of ornaments belonging to $13$ symmetry groups. We have focused on ornaments where motifs do not hint the symmetries of the underling tiling, the pattern formed by repeating shapes. 
As long as an ornament contains sufficient number of motifs (protiles) that are either rotationally asymmetric, or strongly concave, or at least less symmetric than the higher order rotational symmetry of the group, our method works. 
The asymmetry assumption is not a serious restriction. Even in cases where all motifs are symmetric, as long as they do not all centered at the corners of the translational unit, the method still works. If all motifs are symmetric and centered  at the corners of the translational unit, our method can not determine the symmetry group. Nevertheless,  it is still possible to extract a translational repetition lattice. 

As a proof of concept, to show a range of ornaments for which our method works, we have compiled an ornament database of $100$ images. In the set, $14$  of the 
ornaments are images painted by Escher.  All of them are classics, such as the famous Mariposas, Angles and Damons and Lizards. 
The remaining $86$ ornament images are constructed via iOrnament software either by the authors or by several iOrnament artists. 

Because we do not explicitly check for the existence of mirror symmetry, unless indirectly implied by other clues, we sometimes miss mirror reflections. This causes for $3$ groups ($p6m,\,  p4m,\, p3m$) to be classified as belonging to  respective  reflection-less groups of lower symmetry ($p6,\,p4,\, p3$). Nevertheless, 
since their  fundamental domains (double the sizes they should be) contain mirror reflected copies,  recreation of the original patterns using the generation rule of the respective reflection-less groups  is possible. Indeed, this  forms our motivation for postponing the hard mirror reflection checks till they are  implied by indirect clues, such as glide reflections.

\section*{Acknowledgements}
\noindent  The work is funded by TUBITAK Grant 114E204.



\section*{Appendix}
\noindent The Appendix contains the results for the $45$ ornaments that are not included in the main sections of the paper. They are organized according to their symmetry groups and placed in three figures. 

\appendix



\renewcommand{\thefigure}{A\arabic{figure}}

\setcounter{figure}{0}

\begin{figure*}[!htb]
\centering
  \begin{tabular}{m{2.7cm}m{2.7cm}m{2.7cm}m{2.7cm}m{2.7cm}}

\multicolumn{5}{c} {\includegraphics[width=1\linewidth]{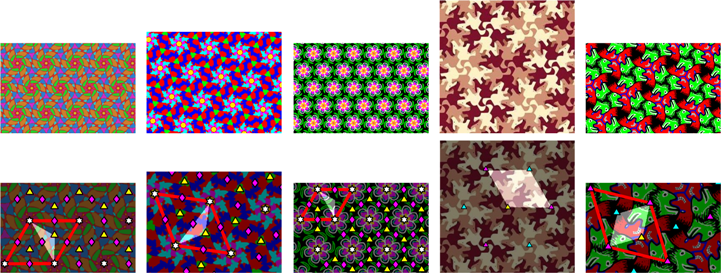} }\\
        \hspace{1.2cm} $p6$& \hspace{1.5cm}$p6$& \hspace{1.7cm}$p6$& \hspace{2cm}$p3$& \hspace{2cm}$p3$\\
\\

\multicolumn{5}{c} {\includegraphics[width=1\linewidth]{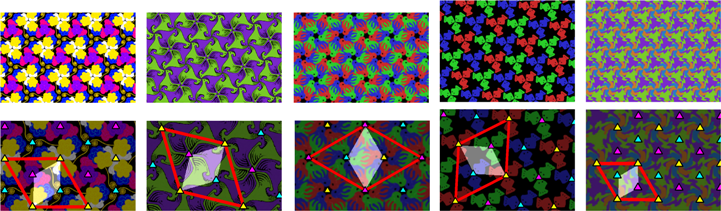} }\\
          \hspace{1.2cm}$p3$& \hspace{1.5cm}$p3$& \hspace{1.7cm}$p3$& \hspace{2cm}$p3$& \hspace{2cm}$p3$\\
\\

\multicolumn{5}{c} {\includegraphics[width=1\linewidth]{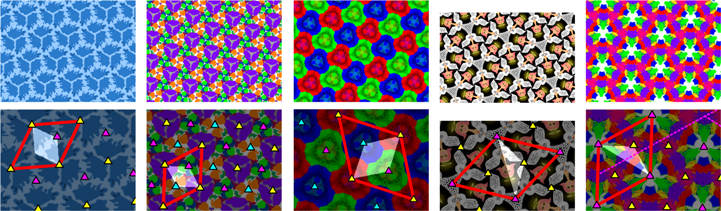} }\\
      \hspace{1.2cm} $p3$& \hspace{1.5cm}$p3$& \hspace{1.5cm}$p3 \tcr{m}$& \hspace{1.7cm}$p31m$& \hspace{1.7cm}$p31m$\\       
   \end{tabular}
\caption{Remaining ornaments - 1}
\label{fig:left}
\end{figure*}

\begin{figure*}[!htb]
\centering
  \begin{tabular}{m{2.7cm}m{2.7cm}m{2.7cm}m{2.7cm}m{2.7cm}}
  
\multicolumn{5}{c} {\includegraphics[width=1\linewidth]{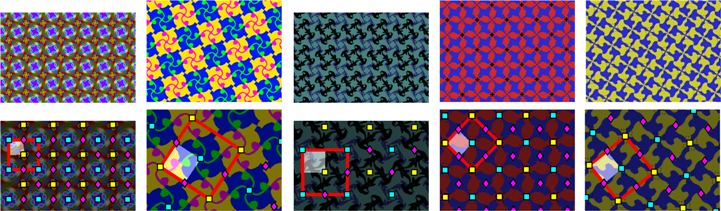} }\\
          \hspace{1.2cm}$p4$& \hspace{1.5cm}$p4$& \hspace{2cm}$p4$& \hspace{2cm}$p4$& \hspace{2cm}$p4$\\
         
        \\ 

\multicolumn{5}{c} {\includegraphics[width=1\linewidth]{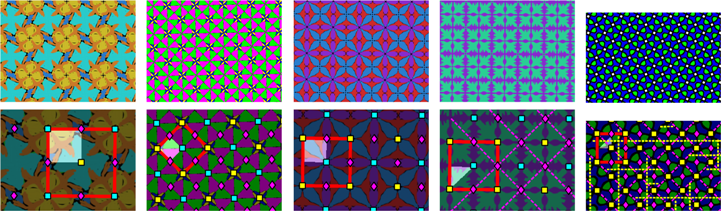} }\\
          \hspace{1.2cm}$p4$& \hspace{1.5cm}$p4$& \hspace{2cm}$p4 \tcr{m}$& \hspace{2cm}$p4m$& \hspace{2cm}$p4g$\\
         \\

\multicolumn{5}{c} {\includegraphics[width=1\linewidth]{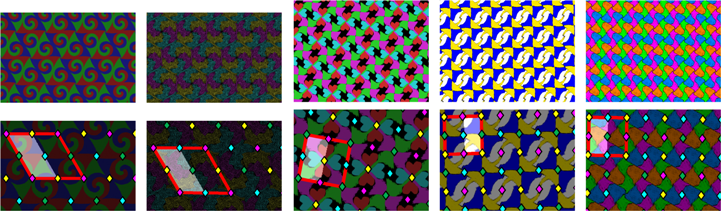} }\\
         \hspace{1.2cm} $p2$& \hspace{1.5cm}$p2$& \hspace{2cm}$p2$& \hspace{2cm}$p2$& \hspace{2cm}$p2$\\

                 \end{tabular}
\caption{Remaining ornaments - 2}
\label{fig:left2}
\end{figure*}

\begin{figure*}[!htb]
\centering
  \begin{tabular}{m{2.7cm}m{2.7cm}m{2.7cm}m{2.7cm}m{2.7cm}}

\multicolumn{5}{c} {\includegraphics[width=1\linewidth]{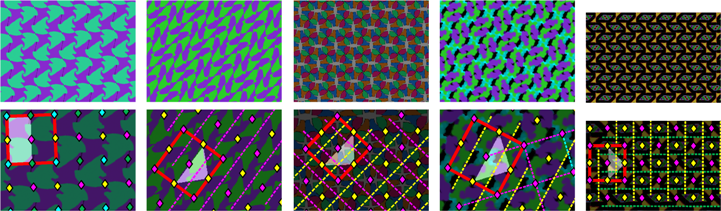} }\\
          \hspace{1.2cm}$p2$& \hspace{1.5cm}$pgg$& \hspace{2cm}$pgg$& \hspace{2cm}$pgg$& \hspace{2cm}$pgg$\\
         \\

           \multicolumn{5}{c} {\includegraphics[width=1\linewidth]{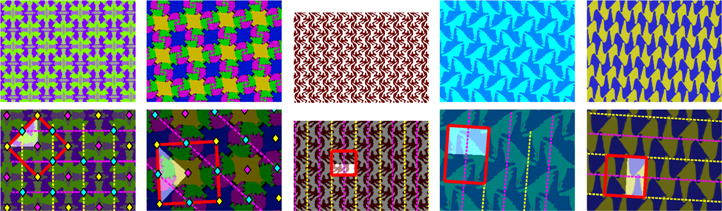} }\\
           \hspace{1.2cm}$cmm$& \hspace{1.5cm}$cmm$& \hspace{2cm}$pg$& \hspace{2cm}$pg$& \hspace{2cm}$pg$\\
\\

\multicolumn{5}{c} {\includegraphics[width=1\linewidth]{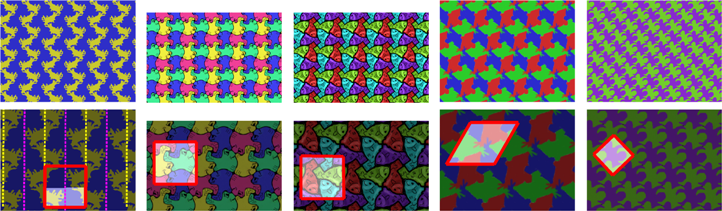} }\\
         \hspace{1.2cm}$pg$& \hspace{1.5cm}$p1$& \hspace{2cm}$p1$& \hspace{2cm}$p1$& \hspace{2cm}$p1$\\

                 \end{tabular}
\caption{Remaining ornaments - 3}
\label{fig:left3}
\end{figure*}

\end{document}